\documentclass{article}
\usepackage{spconf}

\usepackage{amsmath,amssymb,mathtools}
\usepackage{graphicx}
\usepackage{booktabs,multirow}
\usepackage{siunitx}
\usepackage{adjustbox}
\usepackage{url}
\usepackage{pifont}
\usepackage{enumitem}
\setlist{nosep, leftmargin=14pt}

\usepackage{mwe} 

\title{Radiomics-Integrated Deep Learning with Hierarchical Loss for Osteosarcoma Histology Classification}

\name{%
\begin{tabular}{@{}c@{}}
Yaxi Chen$^{1,2}$ \qquad Zi Ye$^{1}$ \qquad Shaheer U.~Saeed$^{2,3,4,5}$\\
Oliver Yu$^{1}$ \qquad Simin Ni$^{6,7}$ \qquad Jie Huang$^{1}$ \qquad Yipeng Hu$^{2,3}$
\end{tabular}%
}

\address{%
$^{1}$Department of Mechanical Engineering, University College London, London, UK\\
$^{2}$UCL Hawkes Institute, University College London, London, UK\\
$^{3}$Department of Medical Physics and Biomedical Engineering, University College London, London, UK\\
$^{4}$School of Engineering and Materials Science, Queen Mary University of London, London, UK\\
$^{5}$Centre for Bioengineering, Queen Mary University of London, London, UK\\
$^{6}$Institute of Orthopaedics and Musculoskeletal Science, University College London, Stanmore, UK\\
$^{7}$Royal National Orthopaedic Hospital, Stanmore, UK%
}

\begin{document}

\maketitle
\begin{abstract}
Osteosarcoma (OS) is an aggressive primary bone malignancy. Accurate histopathological assessment of viable versus non-viable tumor regions after neoadjuvant chemotherapy is critical for prognosis and treatment planning, yet manual evaluation remains labor-intensive, subjective, and prone to inter-observer variability. Recent advances in digital pathology have enabled automated necrosis quantification. Evaluating on test data, independently sampled on patient-level, revealed that the deep learning model performance dropped significantly from the tile-level generalization ability reported in previous studies.
First, this work proposes the use of radiomic features as additional input in model training. We show that, despite that they are derived from the images, such a multimodal input effectively improved the classification performance, in addition to its added benefits in interpretability. Second, this work proposes to optimize two binary classification tasks with hierarchical classes (i.e. tumor-vs-non-tumor and viable-vs-non-viable), as opposed to the alternative ``flat'' three-class classification task (i.e. non-tumor, non-viable tumor, viable tumor), thereby enabling a hierarchical loss. We show that such a hierarchical loss, with trainable weightings between the two tasks, the per-class performance can be improved significantly. Using the TCIA OS Tumor Assessment dataset, we experimentally demonstrate the benefits from each of the proposed new approaches and their combination, setting a what we consider new state-of-the-art performance on this open dataset for this application. Code and trained models: \url{https://github.com/YaxiiC/RadiomicsOS.git}.

\end{abstract}
\section{Introduction}
\label{sec:intro}
Osteosarcoma (OS) is an aggressive primary bone malignancy that predominantly affects children and adolescents \cite{taran2017pediatric}. Accurate histopathological assessment remains essential for determining treatment response and prognosis. In particular, differentiating viable from non-viable OS cells is clinically critical: the proportion of viable tumor after neoadjuvant chemotherapy is a strong independent predictor of patient outcomes. Tumors with fewer than 10\% viable cells are associated with substantially improved survival, whereas those with greater viable fractions exhibit higher recurrence and mortality risk \cite{lee2021osteosarcoma}. Viable tumor cell density directly informs subsequent therapeutic decisions—patients with poor histologic response may receive intensified or alternative regimens, while those with predominantly non-viable tissue can avoid unnecessary toxicity \cite{kawaguchi2024viable}.

Manual evaluation of histopathological slides is labor-intensive, time-consuming, and subject to inter-observer variability, posing challenges for large-scale and objective assessment. Distinguishing viable from non-viable tumor tissue can be particularly difficult due to therapy-induced fibrosis, acellular osteoid, and heterogeneous necrosis, leading to inconsistent grading across observers \cite{picci1985histologic}. As cancer incidence rises and precision oncology expands, there is an urgent need for automated, reproducible, and clinically interpretable diagnostic systems. Early computational pathology pipelines demonstrated the feasibility of automatically segmenting OS whole-slide images (WSIs) into viable tumor, necrotic, and non-tumor compartments, enabling the first steps toward objective quantification of treatment-induced necrosis after neoadjuvant chemotherapy. For instance, several studies applied supervised deep-learning models to replicate pathologists’ necrosis estimation with strong concordance \cite{anisuzzaman2021deep, arunachalam2019viable}. 

Translation of deep learning models into clinical practice remains limited by the inherent rarity and histological heterogeneity of the disease. We argue that it is precisely this low class prevalence and histological variability that challenge current models to generalize, across patients, for simultaneously addressing two distinct classification objectives: histologically pathological vs. healthy and clinically consequential vs. otherwise, in which certain histologically pathological classes, such as non-viable tumors, are not considered clinically significant. From a machine learning perspective, benefits in incorporating these domain-specific prior knowledge in class hierarchy have been demonstrated consistently~\cite{bertinetto2020making, silla2011survey}.
Nevertheless, most existing approaches often treated OS classification as a flat multi-class task, overlooking subtype hierarchies coupled with varying class reverences. Moreover, many reported their preliminary results based on random tile-level data splits, which can lead to information leakage between training and test sets at patient-level \cite{anisuzzaman2021deep,arunachalam2019viable,borji2025advanced,tang2021improving} and, as also shown in our initial investigation, inflation in performance estimates. 

In this study, we propose a multimodal hierarchical learning framework that integrates handcrafted radiomics descriptors, and a structured two-head loss function with learnable uncertainty weighting. The approach aims to improve automated OS histopathology classification by aligning the model design with clinical reasoning, to better support prognostic evaluation and treatment guidance.

\section{Methods}
\label{sec:method}

\subsection{Problem Setting}
Let $\mathcal{X} \subset \mathbb{R}^{H \times W \times 3}$ denote the RGB image space, and $\mathcal{R} \subset \mathbb{R}^{d_r}$ the radiomic feature space.  
Given a dataset
$\mathcal{D} = \{(x_i, y_i)\}_{i=1}^N,$
each sample consists of an image $x_i \in \mathcal{X}$ and its corresponding classification label $y_i \in \mathcal{Y} = \{0, 1, 2\}$, representing 
$\text{Non-tumor}$, $\text{Non-viable-tumor}$, and $\text{viable-tumor}$, respectively.  For each image $x_i$, a predefined radiomic feature vector $r_i \in \mathcal{R}$ is extracted. To encode the pathological hierarchy (Sec.\ref{sec:weighted}), we define two related binary tasks:\[
\setlength{\jot}{2pt} 
\begin{aligned}
y_i^{A} &=
\begin{cases}
0, & y_i = 0, \\[2pt]
1, & y_i \in \{1,2\},
\end{cases}
\quad \text{(Non-tumor vs.\ Tumor)}, 
\\[0.4em]
y_i^{B} &=
\begin{cases}
0, & y_i = 1, \\[2pt]
1, & y_i = 2,
\end{cases}
\quad \text{(Non-viable vs.\ Viable)}.
\end{aligned}
\]
Task~B is defined only for samples that contain tumor tissue, 
i.e., those with labels $y_i \neq 0$, 
whose indices form the set 
$\mathcal{I}_B = \{\, i \mid y_i \neq 0 \,\}$.


\subsection{Multimodal Features and Attention Fusion}
The image encoder 
$f_{\mathrm{img}}(\cdot;\,\boldsymbol{\theta}_{\mathrm{img}}): 
\mathcal{X} \rightarrow \mathbb{R}^{d}$ 
and the radiomic feature encoder 
$f_{\mathrm{rad}}(\cdot;\,\boldsymbol{\theta}_{\mathrm{rad}}): 
\mathcal{R} \rightarrow \mathbb{R}^{d}$ 
are implemented as neural networks parameterized by 
$\boldsymbol{\theta}_{\mathrm{img}}$ and $\boldsymbol{\theta}_{\mathrm{rad}}$, respectively. Both encoders are parameterized and trained jointly end-to-end with 
the following fusion gate and classification heads.

Given an input image–feature pair $(\mathbf{x}, \mathbf{r})$, 
the modality embeddings are computed as
\[
\mathbf{z}_{\mathrm{img}} 
= f_{\mathrm{img}}(\mathbf{x};\,\boldsymbol{\theta}_{\mathrm{img}}) 
\in \mathbb{R}^{d}, 
\qquad
\mathbf{z}_{\mathrm{rad}} 
= f_{\mathrm{rad}}(\mathbf{r};\,\boldsymbol{\theta}_{\mathrm{rad}}) 
\in \mathbb{R}^{d}.
\]
To adaptively combine modalities, we concatenate the image and radiomic embeddings 
and apply a softmax gating network 
$g(\cdot;\,\boldsymbol{\psi}_{g}): \mathbb{R}^{2d} \rightarrow \mathbb{R}^{2}$ 
to obtain normalized attention weights:
\begin{align}
\boldsymbol{\alpha}
&= \mathrm{softmax}\!\bigl(
  g([\mathbf{z}_{\mathrm{img}};\mathbf{z}_{\mathrm{rad}}];\,\boldsymbol{\psi}_{g})
  \bigr)
  \in \mathbb{R}^{2}, \nonumber\\[-2pt]
&\text{s.t. } 
\alpha_{\mathrm{img}} + \alpha_{\mathrm{rad}} = 1,
\quad 
\alpha_{\cdot} \ge 0,
\end{align}
where $g(\cdot;\boldsymbol{\psi}_{g})$ denotes a two-layer MLP with parameters~$\boldsymbol{\psi}_{g}$. 
The fused multimodal representation is then obtained as an 
attention-weighted combination of modality-specific features:
\begin{equation}
\label{eq:fusion}
\mathbf{z}
= \alpha_{\mathrm{img}}\,\mathbf{z}_{\mathrm{img}}
+ \alpha_{\mathrm{rad}}\,\mathbf{z}_{\mathrm{rad}}
\in \mathbb{R}^{d}.
\end{equation}

\subsection{Hierarchical Heads and Per-Head Class Weighting}
\label{sec:weights}
On the fused representation~$\mathbf{z}$, we place two task-specific linear heads: 
a coarse head~(Task~A) for non-tumor vs.\ tumor discrimination, 
and a fine head~(Task~B) for non-viable vs.\ viable tumor classification. 
Each head is implemented as a linear classifier parameterized by 
$\boldsymbol{\omega}_A$ and $\boldsymbol{\omega}_B$, respectively:
\[
h_A(\mathbf{z};\,\boldsymbol{\omega}_A) \in \mathbb{R}^{2}, 
\qquad
h_B(\mathbf{z};\,\boldsymbol{\omega}_B) \in \mathbb{R}^{2}.
\]
The predicted probabilities for each head are denoted by 
$\mathbf{p}_A = [p_{A,0},\,p_{A,1}]^\top$ and 
$\mathbf{p}_B = [p_{B,0},\,p_{B,1}]^\top$, 
representing the per-class probabilities for the coarse and fine tasks, respectively.
The induced three-way distribution $P(c\mid x,r)$ respects the hierarchical 
label structure:

\begin{equation}
\label{eq:hier-prob}
\begin{aligned}
P(c{=}0\mid x,r) &= p_{A,0},\\
P(c{=}1\mid x,r) &= p_{A,1}\, p_{B,0},\\
P(c{=}2\mid x,r) &= p_{A,1}\, p_{B,1}.
\end{aligned}
\end{equation}

We compute per-head class weights
$\mathbf{\beta}^{A}\!\in\!\mathbb{R}_{+}^{2}$ and
$\mathbf{\beta}^{B}\!\in\!\mathbb{R}_{+}^{2}$
from the inverse of the training-set class counts for each head
(coarse: non-tumor vs.~tumor; fine: non-viable vs.~viable).
Weights are normalized to have unit mean and remain fixed during optimization,
serving as static factors in the respective cross-entropy losses.

\begin{figure}[htb]

\begin{minipage}[b]{1.0\linewidth}
  \centering
  \centerline{\includegraphics[width=9.5cm]{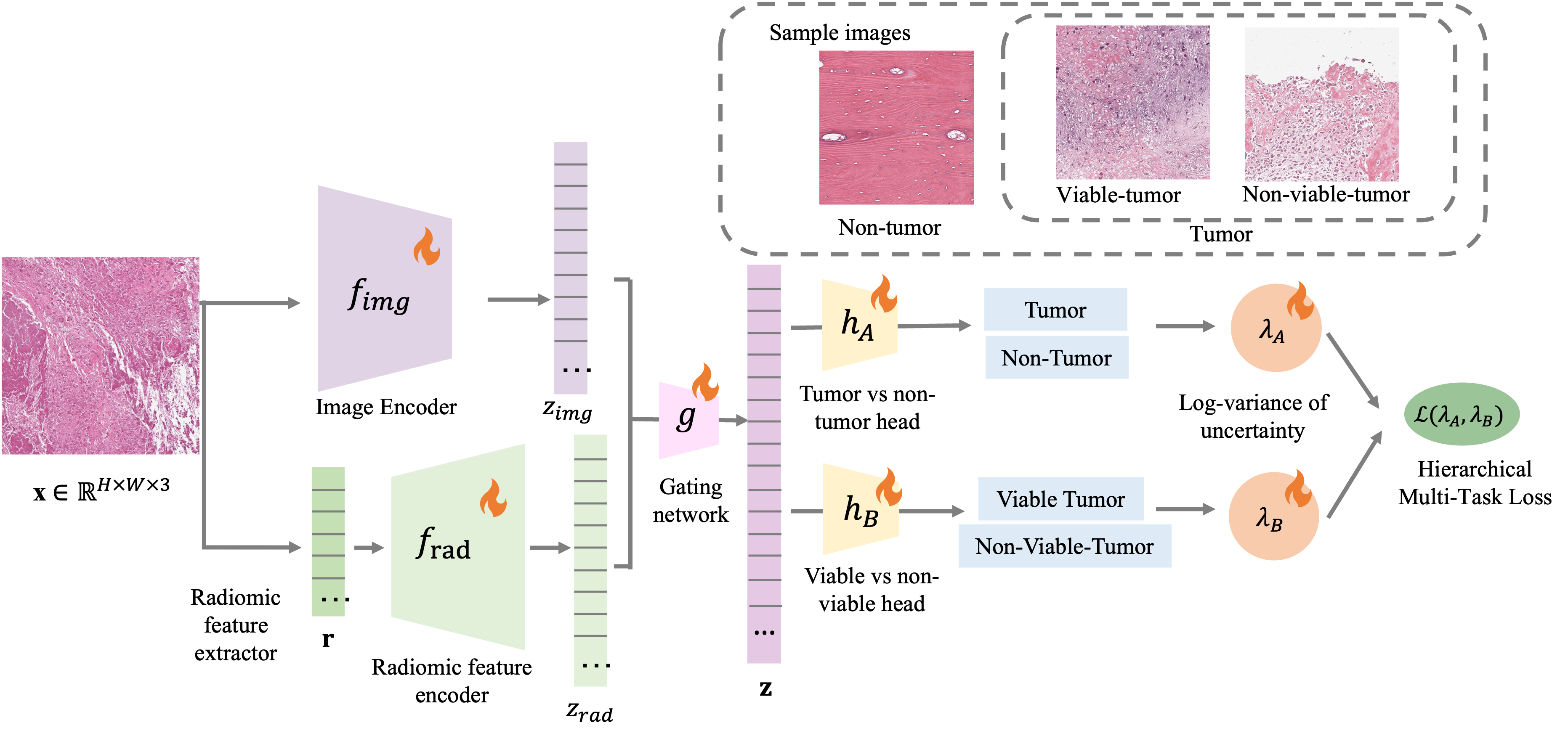}}
  
\end{minipage}
\caption{Overview of the proposed framework for multi-task OS classification.}
\label{fig:res}
\end{figure}

\subsection{Weighted Cross-Entropy for Each Head}
\label{sec:weighted}

Each head is optimized using a weighted cross-entropy loss with its
corresponding class weights (Sec.~\ref{sec:weights}). For a training sample~$i$, the per-sample losses are

\begin{equation}
\label{eq:ce-a}
L_{A,i} = -\, \beta^A_{y_{i}^A} \log p_{A,\,y_{i}^A}.
\end{equation}
for the coarse task (non-tumor vs.\ tumor), and


\begin{equation}
\label{eq:ce-b}
L_{B,i}
= -\, \beta^B_{y_i^B}\, \log p_{B,\,y_i^B},
\quad i \in \mathcal{I}_B,
\end{equation}

for the fine task (non-viable vs.~viable), computed only over tumor samples
$i\!\in\!\mathcal{I}_B$.
Mini-batch losses are obtained by averaging $L_{A, i}$ and $L_{B, i}$
over their respective valid samples.

\subsection{Uncertainty-Weighted Hierarchical Multi-Task Loss}
\label{sec:uncert-loss}
To automatically balance the two tasks, we adopt the \emph{homoscedastic uncertainty weighting} frameworkof Kendall \emph{et al.}~\cite{kendall2018multi}. Homoscedastic (task-dependent) uncertainty is a form of aleatoric noise that is constant across inputs but may differ across tasks. Each task is associated with a
homoscedastic (input-independent) observation variance
$\sigma_A^2$ or $\sigma_B^2$, representing the task-specific noise level. They are treated as learnable scalar parameters and optimized jointly with all network weights during training to adaptively balance the two task losses.

Following the probabilistic derivation in~\cite{kendall2018multi},
the negative log-likelihood introduces weighting factors
$1/\sigma_A^2$ and $1/\sigma_B^2$ for the two tasks,
along with regularization terms in $\log\sigma_A^2$ and $\log\sigma_B^2$.
For numerical stability, we optimize their logarithms,
$\lambda_A = \log\sigma_A^2$ and $\lambda_B = \log\sigma_B^2$,
which ensure positivity and yield learnable weights
$e^{-\lambda_A}$ and $e^{-\lambda_B}$.

The joint objective over the two tasks is
\begin{equation}
\label{eq:uncertainty-loss}
\mathcal{L}(\lambda_A,\lambda_B)
= e^{-\lambda_A}\,\overline{L}_A
+ e^{-\lambda_B}\,\overline{L}_B
+ \eta\,(\lambda_A+\lambda_B),
\end{equation}
where $\overline{L}_A$ and $\overline{L}_B$ are the mean
weighted cross-entropies (Eqs.~\ref{eq:ce-a}–\ref{eq:ce-b}),
and $\eta>0$ is a small regularization constant prevents the task uncertainties from growing unbounded.
Intuitively, tasks with larger uncertainty
($\sigma_A$ or $\sigma_B$) receive smaller effective weights
($e^{-\lambda_A}$ or $e^{-\lambda_B}$),
allowing the model to down-weight more uncertain tasks.

\subsection{Training and Optimization}
All components of the proposed framework, including the image encoder 
$f_{\mathrm{img}}(\cdot;\,\boldsymbol{\theta}_{\mathrm{img}})$, the radiomic encoder 
$f_{\mathrm{rad}}(\cdot;\,\boldsymbol{\theta}_{\mathrm{rad}})$, the fusion gate $g(\cdot;\,\boldsymbol{\psi}_{g})$, 
the classification heads $h_A(\cdot;\,\boldsymbol{\omega}_A)$ and $h_B(\cdot;\,\boldsymbol{\omega}_B)$, and the uncertainty parameters 
$\lambda_A$ and $\lambda_B$, are optimized jointly in an end-to-end manner. 
The overall training objective is expressed as
\begin{equation}
\label{eq:final-opt}
(\boldsymbol{\theta}_{\mathrm{img}}^*,\,\boldsymbol{\theta}_{\mathrm{rad}}^*,\,
\boldsymbol{\psi}_{g}^*,\,\boldsymbol{\omega}_A^*,\,\boldsymbol{\omega}_B^*,\,
\lambda_A^*,\,\lambda_B^*)
= \arg\min
\mathcal{L}(\lambda_A,\lambda_B),
\end{equation}

where $\mathcal{L}(\lambda_A,\lambda_B)$ is the 
uncertainty-weighted hierarchical loss defined in 
Eq.~(\ref{eq:uncertainty-loss}). Equation~(\ref{eq:final-opt}) is optimized via gradient descent using the 
AdamW optimizer, jointly updating all network and uncertainty parameters.


\section{Experiments and Results}
\label{sec:pagestyle}
\begin{table}[!t]
  \centering
  \caption{Summary of overall classification metrics across model backbones, hierarchical loss configurations, and inclusion of radiomics features.}
  \label{tab:sum}
 
  \setlength{\tabcolsep}{5pt}          
  \renewcommand{\arraystretch}{1.05}   
  \scriptsize
  \begin{adjustbox}{max width=\textwidth}
  \begin{tabular}{l c c l cccc}
    \toprule
    \multirow{2}{*}{Backbone} & \multirow{2}{*}{H-loss} & \multirow{2}{*}{Rad.} &  \multicolumn{4}{c}{Overall summary} \\
    \cmidrule(lr){4-7}
     & & & Acc. & $F_{1,\text{macro}}$ & $F_{1,\text{weighted}}$ & $\mathrm{AUC}_{\text{ovr}}$ \\
    \midrule

    IncV3(3-class) & \ding{55} & \ding{55} & 0.73$\pm$0.02 & 0.73$\pm$0.02 & 0.73$\pm$0.03 & 0.90$\pm$0.00 \\
     \midrule
     
    ViT & \ding{55} & \ding{55}  & 0.71$\pm$0.01 & 0.71$\pm$0.01 & 0.70$\pm$0.01 & 0.87$\pm$0.00 \\
    \midrule

    EffNet & \ding{55} & \ding{55} &  0.65$\pm$0.01 & 0.56$\pm$0.12 & 0.67$\pm$0.05 & 0.84$\pm$0.04 \\
    \midrule

    
    IncV3 & \ding{55} & \ding{55} & 0.73$\pm$0.00 & 0.74$\pm$0.02 & 0.73$\pm$0.00 & 0.90$\pm$0.01 \\
     \midrule
    IncV3 & \ding{55} & \ding{51} &  0.80$\pm$0.03 & 0.80$\pm$0.02 & 0.80$\pm$0.02 & 0.93$\pm$0.01 \\
     \midrule
    IncV3 & \ding{51} & \ding{55} &  0.75$\pm$0.02 & 0.75$\pm$0.02 & 0.75$\pm$0.02 & 0.92$\pm$0.00 \\
     \midrule

    IncV3 \textbf{(Ours)} & \ding{51} & \ding{51} &  \textbf{0.86$\pm$0.01} & \textbf{0.86$\pm$0.01} & \textbf{0.86$\pm$0.01} & \textbf{0.96$\pm$0.00} \\
    \bottomrule
  \end{tabular}
  \end{adjustbox}
\end{table}

\begin{table}[!t]
  \centering
  \caption{Per-class performance metrics for each backbone, with and without hierarchical loss and radiomics features.}
  \label{tab:perclass}
  \setlength{\tabcolsep}{3pt}          
  \renewcommand{\arraystretch}{0.95}   
  \scriptsize                         
  \begin{adjustbox}{max width=\textwidth, max totalheight=\textheight, keepaspectratio}
  \begin{tabular}{l c c l cccc}
    \toprule
    \multirow{2}{*}{Backbone} & \multirow{2}{*}{H-loss} & \multirow{2}{*}{Rad} & \multirow{2}{*}{type} & \multicolumn{4}{c}{Per-class} \\
    \cmidrule(lr){5-8}
     &  &  &  & Sen@Spe90 & Spe@Sen90 & $F_{1}$ & AUC \\
        \midrule
    \multirow{3}{*}{IncV3(3-class)}
      & \multirow{3}{*}{\ding{55}} & \multirow{3}{*}{\ding{55}}
        & NT        & 0.78±0.02 & 0.52±0.06   & 0.73±0.02   & 0.89±0.00   \\
      &                    & 
        & NVT & 0.71±0.03 & 0.72±0.04   & 0.69±0.03   & 0.89±0.01   \\
      &                    & 
        & VT     & 0.79±0.03 & 0.68±0.05   & 0.78±0.03   & 0.91±0.01   \\
         \midrule
    
    \multirow{3}{*}{ViT}
      & \multirow{3}{*}{\ding{55}} & \multirow{3}{*}{\ding{55}}
        & NT        & 0.75±0.01    & 0.41±0.03   & 0.74±0.01   & 0.88±0.00   \\
      &                    & 
        & NVT & 0.58±0.04    & 0.50±0.04   & 0.67±0.02   & 0.85±0.00   \\
      &                    & 
        & VT     & 0.66±0.04    & 0.70±0.04   & 0.67±0.02   & 0.89±0.01   \\

    \midrule
    \multirow{3}{*}{EffNet}
      & \multirow{3}{*}{\ding{55}} & \multirow{3}{*}{\ding{55}}
        & NT        & 0.72±0.15    & 0.65±0.09   & 0.74±0.04   & 0.90±0.05   \\
      &                    & 
        & NVT & 0.53±0.08    & 0.60±0.14   & 0.59±0.14   & 0.84±0.03   \\
      &                    & 
        & VT     & 0.24±0.34    & 0.64±0.04   & 0.36±0.46   & 0.79±0.08   \\
    \midrule
    
    \multirow{3}{*}{IncV3}
          & \multirow{3}{*}{\ding{55}} & \multirow{3}{*}{\ding{55}}
        
        & NT        & 0.80±0.02    & 0.78±0.03   & 0.79±0.04   & 0.93±0.01   \\
      &                    & 
        & NVT & 0.67±0.08    & 0.65±0.07   & 0.70±0.03   & 0.87±0.05   \\
      &                    & 
        & VT     & 0.67±0.08    & 0.69±0.07   & 0.72±0.02   & 0.89±0.02   \\
          \midrule

    \multirow{3}{*}{IncV3}
      & \multirow{3}{*}{\ding{55}} & \multirow{3}{*}{\ding{51}}
        & NT        & 0.81±0.06    & 0.77±0.07   & 0.78±0.03   & 0.92±0.00   \\
      &                    & 
        & NVT & 0.78±0.05    & 0.80±0.00   & 0.78±0.05   & 0.93±0.05   \\
      &                    & 
        & VT     & 0.87±0.07    & 0.87±0.05   & 0.83±0.01   & 0.95±0.01   \\
          \midrule
      \multirow{3}{*}{IncV3}
      & \multirow{3}{*}{\ding{51}} & \multirow{3}{*}{\ding{55}}
        & NT        & 0.82±0.02 & 0.73±0.04   & 0.74±0.01   & 0.93±0.01   \\
      &                    & 
        & NVT & 0.69±0.01 & 0.60±0.04   & 0.71±0.05   & 0.88±0.00   \\
      &                    & 
        & VT     & 0.88±0.03 & 0.87±0.04   & 0.81±0.00   & 0.96±0.00   \\
         \midrule
      \multirow{3}{*}{IncV3 \textbf{(Ours)}}
      & \multirow{3}{*}{\ding{51}} & \multirow{3}{*}{\ding{51}}
        & NT        & 0.92±0.02    & 0.91±0.02   & 0.84±0.00   & 0.97±0.00   \\
      &                    & 
        & NVT & 0.83±0.01    & 0.80±0.02   & 0.84±0.01   & 0.93±0.00   \\
      &                    & 
        & VT    & 0.95±0.01    & 0.94±0.00   & 0.89±0.01   & 0.97±0.00   \\

    \bottomrule
  \end{tabular}
  \end{adjustbox}
\end{table}

\subsection{Dataset and Implementation Details}
\label{subsec:dataset}

We used the publicly available OS tumor assessment dataset from The Cancer Imaging Archive (TCIA)~\cite{leavey2019osteosarcoma}.
The collection comprises 1,144 H\&E-stained histopathology tiles at $10\times$ magnification from surgical resection specimens of OS patients (1995--2015). 
All tiles were pathologist-annotated into three histologic categories:
\emph{Non-tumor} (536 tiles), 
\emph{Non-viable-tumor} (263 tiles), 
and \emph{viable-tumor} (345 tiles). For preprocessing, each tile was loaded in RGB format, resized to $224\times224$ pixels (for EfficientNet-B0) and normalized using ImageNet mean and standard deviation. 
To augment the training set, random horizontal flips and small rotations ($\pm15^\circ$) were applied.

Previous studies using the same TCIA OS Tumor Assessment dataset have typically adopted random splits for training and testing. However, since multiple tiles originate from the same patient specimen, such random partitioning can inadvertently introduce data leakage, leading to overly optimistic performance estimates ~\cite{anisuzzaman2021deep, arunachalam2019viable, borji2025advanced, tang2021improving}.
In this study, we aim to assess generalization ability over subjects, and therefore employ a patient-level split, ensuring that no image tiles from the same patient appear across different subsets.

We extract features using a PyTorch implementation based on the PyRadiomics library~\cite{van2017computational}. To maintain interpretability and reduce redundancy, only first-order and 2D shape features were extracted, resulting in a total of 29 feature types. All features were standardized using training-set mean/std and applied to validation/test sets. The image encoder $f_{\mathrm{img}}$ was instantiated with a convolutional or transformer backbone followed by a linear projection. 
The radiomic encoder $f_{\mathrm{rad}}$ consisted of a two-layer MLP. 
The fusion gate $g$ was a two-layer MLP followed by a softmax. 
All models were implemented in PyTorch~2.5.1 and trained on NVIDIA Quadro GV100 GPUs (32~GB VRAM).  
We used the AdamW optimizer with an initial learning rate of $1\times10^{-4}$, 
weight decay $1\times10^{-4}$, and batch size 16. For all experiments, the regularization constant in Eq.~(\ref{eq:uncertainty-loss}) 
was fixed to $\eta = 0.2$

At inference, we compute $P(c \mid x, r)$ according to~\eqref{eq:hier-prob} and evaluate performance using accuracy, macro/weighted F1-scores, and one-vs-rest (OvR) macro-AUC. 
To provide class-level insights, we also report per-class controlled sensitivity/specificity, F1-score, and AUC. All metrics were averaged over five independent runs.

\subsection{Comparison and ablation studies}
We conducted a series of ablation experiments to assess the contribution of (1) the hierarchical loss, (2) radiomic feature integration, and (3) different backbone architectures. 
Among the evaluated architectures, InceptionV3 consistently outperformed EfficientNet-B0 and ViT. Incorporating the hierarchical loss improved performance, particularly for InceptionV3, macro-F1 increased from 0.74 to 0.75 and AUC from 0.90 to 0.92 (Table~\ref{tab:sum}), confirming that modeling coarse-to-fine dependencies enhances class discrimination.
Adding radiomic features further boosted accuracy and AUC (e.g., 0.80 vs.\ 0.75 macro-F1 for InceptionV3), highlighting the complementary nature of handcrafted shape and intensity descriptors.
The proposed \textbf{InceptionV3 + hierarchical loss + radiomics} configuration achieved the best overall performance, with accuracy significantly higher than all other models (with all $p$-values~$<0.05$). Per-class results (Table~\ref{tab:perclass}) show that the full model improved non-viable tumor recognition, with Sen@Spe90 rising from 0.69 to 0.83 and AUC from 0.87 to 0.93, while maintaining strong non-tumor (Sen@Spe90~0.92, AUC~0.97) and viable-tumor performance (Sen@Spe90~0.95, AUC~0.97). 
These results confirm that hierarchical supervision and multimodal fusion together enhance discrimination across all clinically relevant tissue types.

Per-class results (Table~\ref{tab:perclass}) show that the proposed full model substantially improved non-viable tumor recognition, for the non-viable-tumor class, Sen@Spe90 was significantly higher than all other configurations with all $p$-values~$<0.05$. For the viable-tumor class, both Sen@Spe90 and Spe@Sen90 showed significant improvements over competing models, with all $p$-values~$<0.05$. 
These results demonstrate that hierarchical supervision combined with multimodal fusion enhances discrimination across all clinically relevant tissue types.

\section{Conclusion \& Discussion}

We proposed a multimodal hierarchical learning framework for osteosarcoma histopathology classification that integrates deeplearning image features with radiomic descriptors, in addition to a new uncertainty-weighted multi-task supervision. Experiments on the TCIA osteosarcoma dataset demonstrate that combining hierarchical loss and radiomics consistently improves classification performance, particularly for the challenging non-viable tumor subtype. The best-performing configuration (\textbf{InceptionV3 + hierarchical loss + radiomics}) achieved \textbf{0.83 accuracy} and \textbf{0.94 AUC} under patient-level evaluation, outperforming single-modality or non-hierarchical baselines.

Beyond performance, the use of radiomic features introduces intrinsic interpretability through explicit shape descriptors, complementing the implicit representations learned by deep networks. The hierarchical formulation further supports clinically meaningful reasoning across tumor progression stages. Future extensions may leverage recent advances in radiomics interpretability~\cite{chen2025patient,chen2025radiomic} to enhance transparency and clinical trust.



\bibliographystyle{IEEEbib}
\bibliography{strings}

\end{document}